\documentclass[sn-standardnature]{sn-jnl}

\usepackage[utf8]{inputenc}
\usepackage[T1]{fontenc}
\usepackage{amsmath, amssymb, amsthm}
\usepackage{graphicx}
\usepackage{geometry}
\usepackage{lmodern}
\usepackage{microtype}
\usepackage{booktabs}
\usepackage{url}
\usepackage{algorithm}
\usepackage{algorithmic}
\usepackage{setspace}
\usepackage{caption}
\usepackage{xcolor}

\title{RIS-Kernel: A Model-Agnostic Architecture for Long-Context LLM Inference via Sparse Attention}

\author*[1]{\fnm{Anderson R.} \sur{Santos}}\email{santosardr@ufu.br}
\affil*[1]{\orgdiv{Faculty of Computing}, \orgname{Federal University of Uberlândia (UFU)}, \country{Brazil}}

\begin{document}

\abstract{
Full self-attention in large language models scales as $O(N^2)$, which limits long-context document analysis to 65,536 tokens and requires costly GPU clusters. The Reduced Interaction Sampling (RIS) inference engine addresses this constraint as a model-agnostic architecture. Without modifying weights, RIS introduces sparse stochastic attention geometries that operate under extreme connection pruning (pruning up to 99\% of pairwise interactions under 1\% density) while preserving factual retrieval. While foundational graph-theoretic RIS formulations establish an asymptotic $O(N \log N)$ interaction bound under logarithmic edge budgets, this paper evaluates a practical fixed-density sampling regime on unaccelerated commodity CPU systems. We validate RIS on Qwen2-1.5B-Instruct across two regimes. In controlled evaluations at 32,768 tokens (where native dense attention serves as the upper bound), RIS-Stochastic at 1\% density and 70 ensemble seeds achieves 75.00\% accuracy, outperforming the native dense baseline (71.88\%), while RIS-Stochastic at 5\% density and 10 seeds matches it (71.88\%). This demonstrates that sparse attention acts as a regularizer: low density (1\%) over multiple seeds filters out sequence-level noise, whereas higher density (5\%) reintroduces distractor noise. Crucially, a controlled ablation experiment evaluating a single static random mask at matched budget ($N=1, d=0.50$, sampling 50\% of pairwise connections) yields only 53.12\% accuracy, proving that multi-seed topological diversity, rather than raw edge density, underpins contextual recovery. Under the tightest budget, RIS-Structural reaches 65.62\% accuracy at 1\% effective density with 10 seeds (recovering 50.0\% of the contextual gap relative to the 59.38\% zero-context floor) and reaches 68.75\% (75.0\% gap recovery) with 30 seeds. At 65,536 tokens, where dense attention triggers out-of-memory faults, RIS yields retrieval gains of up to 14.06 percentage points over the zero-context floor (51.56\%), which is confirmed as marginally significant under McNemar's paired test ($p = 0.078 < 0.10$). All evaluations run on commodity, unaccelerated CPU servers (16--128\,GB of RAM), demonstrating that long-context LLM inference is feasible on standard academic hardware without GPU acceleration.
}


\maketitle

\section{Introduction}

Processing large document corpora through large language models requires hardware infrastructure that few research groups possess \cite{dao2022flashattention}. Full self-attention scales as $O(N^2)$ \cite{vaswani2017attention}. At 65,536 tokens, inference costs confine deep textual analysis to institutions operating large GPU clusters. The bottleneck is algorithmic; its practical consequence is a dependency on hardware capital that most groups cannot meet.

In the foundational work~\cite{santos2024ris}, we established the theoretical foundation of Reduced Interaction Sampling (RIS), proving mathematically that the $O(N^2)$ attention bottleneck can be bypassed via stochastic sparsification \cite{child2019generating,zaheer2020bigbird} while preserving factual retrieval. The present paper validates this mechanism empirically under LLM inference. Turning this theory into a functional inference engine required solving three engineering problems: (1) generating sparse geometric masks in $O(N^2)$ space using lightweight boolean structures to avoid out-of-memory faults during allocation; (2) normalizing stochastic token fragments without diluting their competitive weight; and (3) preventing the collapse of positional encodings when extending windows far beyond native training limits. This paper describes the RIS-Kernel architecture, a systems-level implementation that injects runtime sparsity directly into unmodified language models \cite{li2023adapter} through two sampling regimes: \textit{Stochastic Mode} for global coverage and \textit{Structural Mode} for local community preservation.

Validation focuses on Qwen2-1.5B-Instruct~\cite{qwen2025} and TinyLlama-1.1B~\cite{zhang2024tinyllama}. Sub-2B parameter models define the most restrictive regime for long-context factual retrieval: compact enough to run on CPU hardware without GPU acceleration, yet architecturally complete enough to respond meaningfully to context delivered at inference time.

Factual accuracy demands inference-time context delivery \cite{lewis2020rag,ram2023incontext}. RIS enables this on unaccelerated CPU hardware. While asymptotic $O(N \log N)$ complexity describes the graph-theoretic connection threshold, RIS serves here as an uncompiled reference architecture that prunes up to 99\% of pairwise interactions in standard PyTorch attention masks, enabling high-memory commodity CPU servers to process \cite{chen2023efficientcpu} 65k-token contexts without GPU hardware.

\section{Results}

\subsection{Hardware Profiling and Reproducibility Baseline}

Experiments were conducted on two unaccelerated CPU servers, without GPU support. The 64k scalability benchmarks (Section~\ref{sec:exp_b}) ran on a high-memory Xeon workstation (\textbf{ibteci}: 2 sockets, 20 physical cores / 40 threads total, 128\,GB DDR4). The primary 32k controlled evaluation (Section~\ref{sec:exp_a}) --- including the full dense $O(N^2)$ baseline and all RIS hyperparameter sweeps --- was mostly executed on a desktop-grade server (\textbf{bioinfo}: Intel Core i7-3770, 4 physical cores / 8 threads, 16\,GB DDR4), representative of standard consumer hardware repurposed as institutional computing nodes. Some extreme density configurations were executed on the \textbf{ibteci} server due to out-of-memory errors on the \textbf{bioinfo} machine. The sparsity structure is hardware-agnostic, so all results reported here carry over directly to GPU-accelerated deployments at lower latency.

The dominant bottleneck on CPU \cite{wulf1995hitting,williams2009roofline} is L3 cache saturation. For a 65k-token sequence,
\texttt{matmul\_qk} (score computation) takes $\sim$19\,s; \texttt{matmul\_av} (value aggregation)
takes 350\,s. The 18$\times$ gap reflects data-movement cost from DDR to cache \cite{ivanov2021data}, not arithmetic.
Float16 emulation \cite{intel2023optimization} on CPUs lacking native AVX-512 BF16 pushed single optimization steps past
60 hours; Float32 restored predictable runtimes. A standard 65k-token prefill occupies
approximately 38\,GB under the 40-seed RIS ensemble configuration. Larger ensemble configurations
approach the 128\,GB physical ceiling, bounding the hyperparameter sweep.
Thread-count discipline is a hard constraint: unconstrained PyTorch scheduling (101 threads) saturated the memory bus \cite{williams2009roofline} and extended prefill time by a factor of 6$\times$. Thread allocation was restricted near physical core boundaries --- 4 threads on the 4-core bioinfo server and 8 threads on the multi-socket ibteci server --- which restored optimal throughput.

\subsection{Empirical Evaluation}
\label{sec:empirical_eval}

The evaluation contexts for the quantitative benchmarks are constructed by concatenating four scientific manuscripts covering distinct biological domains: genomic characterization of \textit{Acetilactobacillus jinshanensis} (ajinshanensis), anaerobic marine methane oxidation (aom), protein-protein interaction network prediction (genppi), and Jatai bee larval food metagenomics (meta). Prior to concatenation, all metadata, title pages, keywords, and abstracts preceding the introduction are removed.

\subsubsection{Experiment A: Controlled Precision Comparison (32k Window)}
\label{sec:exp_a}

At Qwen2's native positional limit of 32,768 tokens, full dense attention is computationally feasible and serves as the ground-truth upper bound. Using the balanced 32-question set, the zero-context floor ($w=0$) is 59.38\% and the native dense target is 71.88\%, a gap of 12.5 percentage points. The goal of this experiment is to measure how much of that gap RIS-Kernel recovers.

Table~\ref{tab:results_32k} compiles accuracy and context recovery rates across a grid of attention densities (1\%, 2\%, 5\%) and ensemble seeds (1 to 100) for both Stochastic and Structural modes.

RIS-Stochastic at 1\% density and 70--80 seeds achieves 75.00\% accuracy, exceeding the native dense baseline of 71.88\% and yielding a 125.0\% context recovery rate. At this extreme sparsity level, the noise pruned by sparsification outweighs the information lost to it: the ensemble acts as an attention regularizer. RIS-Stochastic also matches the dense baseline exactly (71.88\%, 100\% recovery) at multiple points in the density--seed space: 1\% density at 60, 90, and 100 seeds; 2\% density at 40, 50, and 70 seeds; and 5\% density at 10 seeds.

\begin{figure}[htbp]
    \centering
    \includegraphics[width=0.85\linewidth]{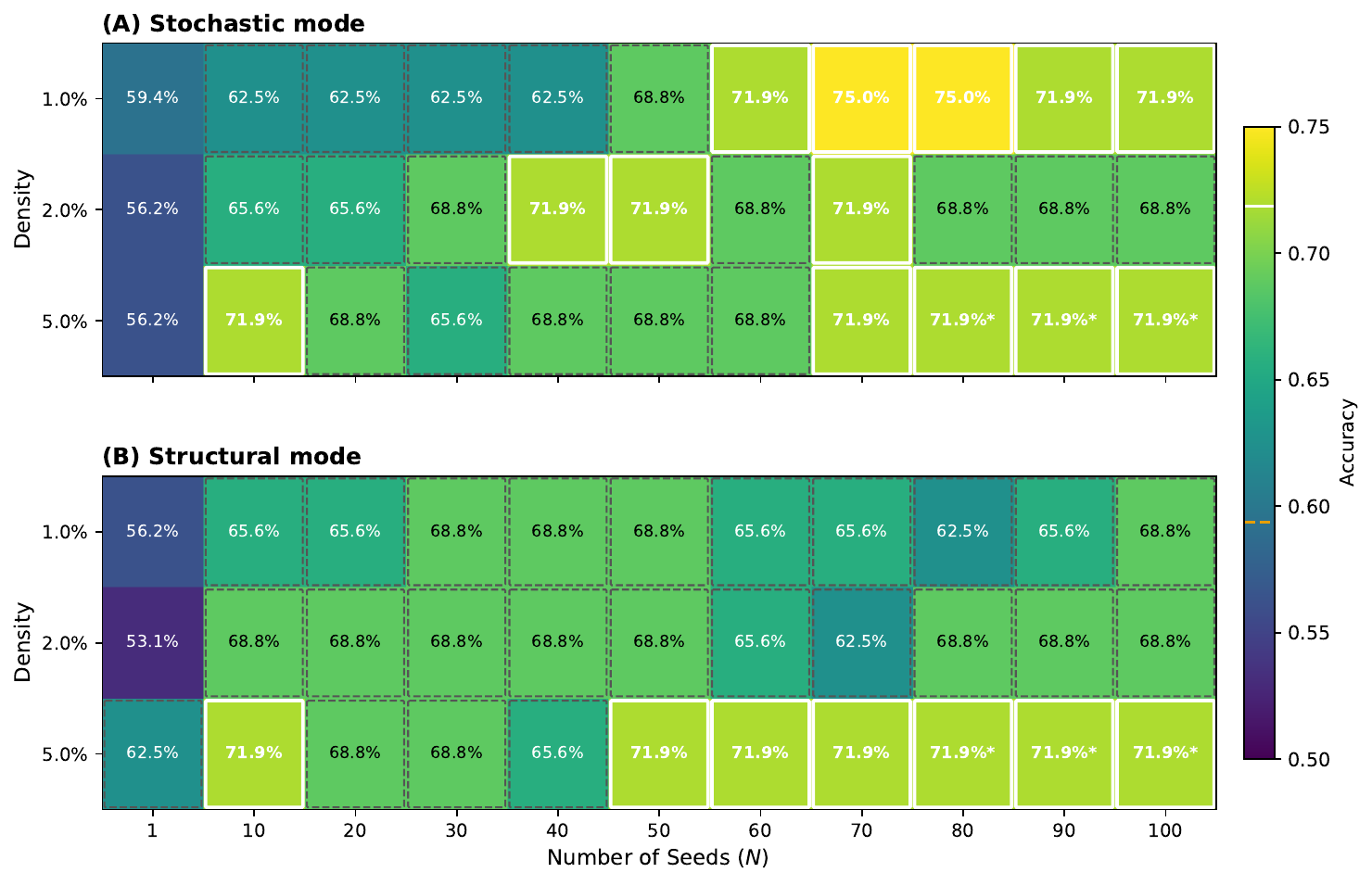}
    \caption{Performance surface of RIS on the balanced 32k window for (a) RIS-Stochastic and (b) RIS-Structural modes. Cells highlighted with blue borders match or exceed the native dense baseline of 71.88\% (seeds/density configurations that fully recover contextual information). Dashed borders show configurations that outperform the zero-context baseline of 59.38\%. Cells marked with an asterisk (*) represent configurations executed on the high-memory \textit{ibteci} server due to out-of-memory (OOM) errors on the desktop-grade \textit{bioinfo} server.}
    \label{fig:heatmap_32k}
\end{figure}

\begin{figure}[htbp]
    \centering
    \includegraphics[width=0.85\linewidth]{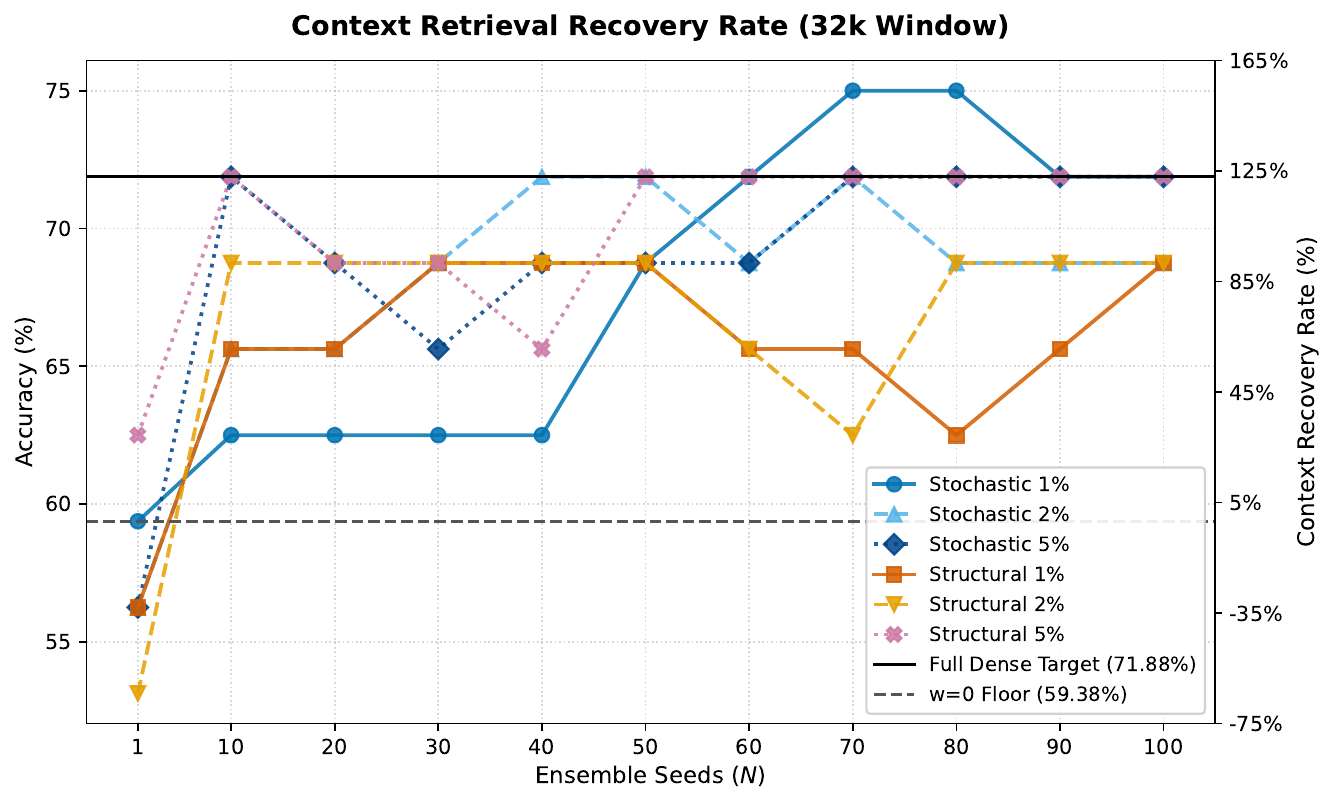}
    \caption{Ensemble scaling and context recovery rate on the balanced 32k window. The left y-axis shows absolute accuracy; the right y-axis indicates the percentage of the full-attention gap recovered relative to the $w=0$ baseline. Under Stochastic 5\% (10 seeds), the RIS architecture matches the native dense target (71.88\%) exactly. Under Structural 1\% (10 seeds), the model recovers 75\% of the contextual gap at 100$\times$ computational sparsity.}
    \label{fig:context_recovery_32k}
\end{figure}

\begin{figure}[htbp]
    \centering
    \includegraphics[width=0.85\linewidth]{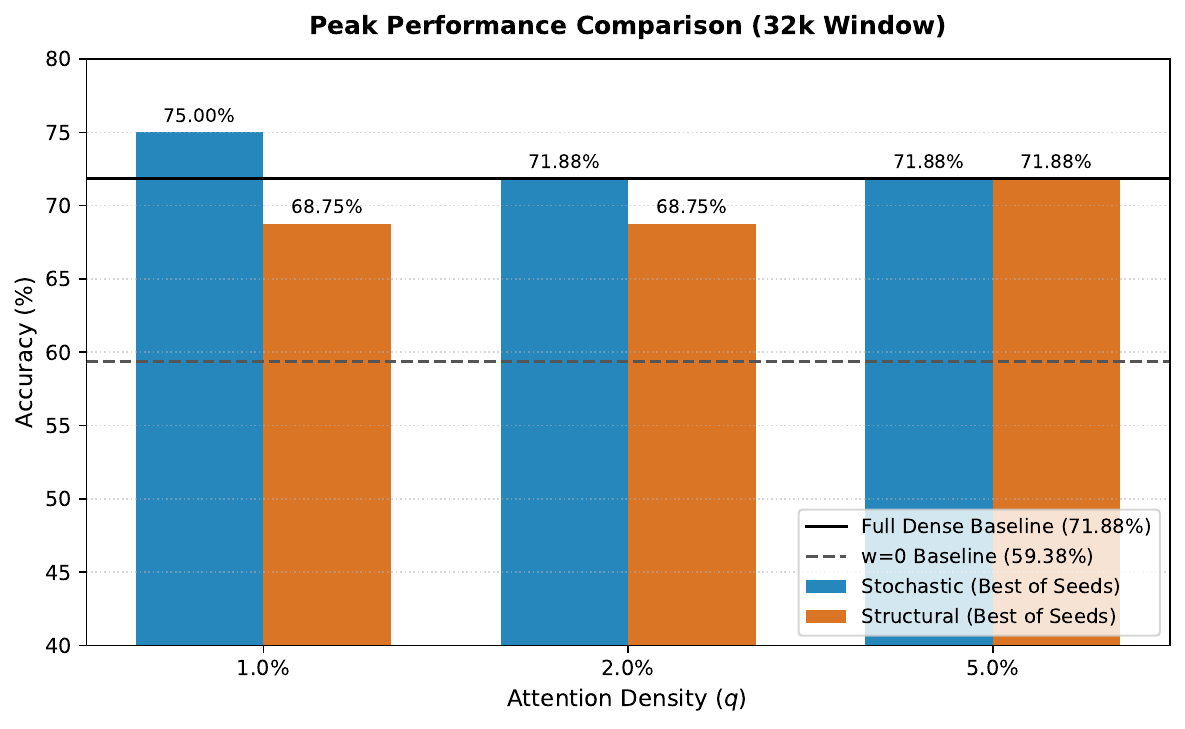}
    \caption{Peak accuracy comparison by sampling mode across attention densities (1\%, 2\%, 5\%) on the balanced 32k window. Horizontal reference lines indicate the zero-context floor (59.38\%) and the full-attention native dense target (71.88\%). Both modes converge to the dense target as density increases; the Structural mode reaches that convergence at substantially lower seed counts when density is at 1\%.}
    \label{fig:mode_comparison_32k}
\end{figure}

At low density, the Structural mode converges to the dense baseline with fewer seeds. Under a constrained budget --- 1\% effective density and 10 seeds --- RIS-Structural achieves 65.62\% accuracy (recovering 50.0\% of the contextual gap relative to the 59.38\% floor), outperforming the Stochastic mode at the same budget (62.50\%). Increasing seeds to 30 brings accuracy to 68.75\% (75.0\% gap recovery), matching the performance reached by the Stochastic mode at 50 seeds. At 2\% effective density, RIS-Structural attains 68.75\% already at 10 seeds, and reaches the native dense target (71.88\%) at 30 seeds versus 40 for the Stochastic mode. The block-clique geometry preferentially captures distal anchors that uniform random sampling misses at low density. Figures~\ref{fig:heatmap_32k}--\ref{fig:mode_comparison_32k} map these results in detail.

\subsubsection{Experiment B: Scalability and Extrapolation (64k Window)}
\label{sec:exp_b}

At 65,536 tokens --- a $2\times$ extrapolation beyond Qwen2's native limit --- a native dense baseline is physically prohibited by memory requirements on standard Xeon testbeds, producing OOM faults. Scalability is evaluated using the expanded 64-question set, with a zero-context floor of 51.56\%.

Results diverge sharply with the RoPE scaling method (detailed in Section~\ref{sec:rope}). Under linear interpolation, accuracy collapses to 15.6\% at 1\% density and 23.4\% at 5\% density with a single seed --- at or below the 20\% expected from random guessing. The stochastic ensemble partially compensates through ensemble coverage: with 40 seeds at 5\% density, accuracy recovers to 59.4\%, above the zero-context baseline.

\begin{figure}[htbp]
    \centering
    \includegraphics[width=1.0\textwidth]{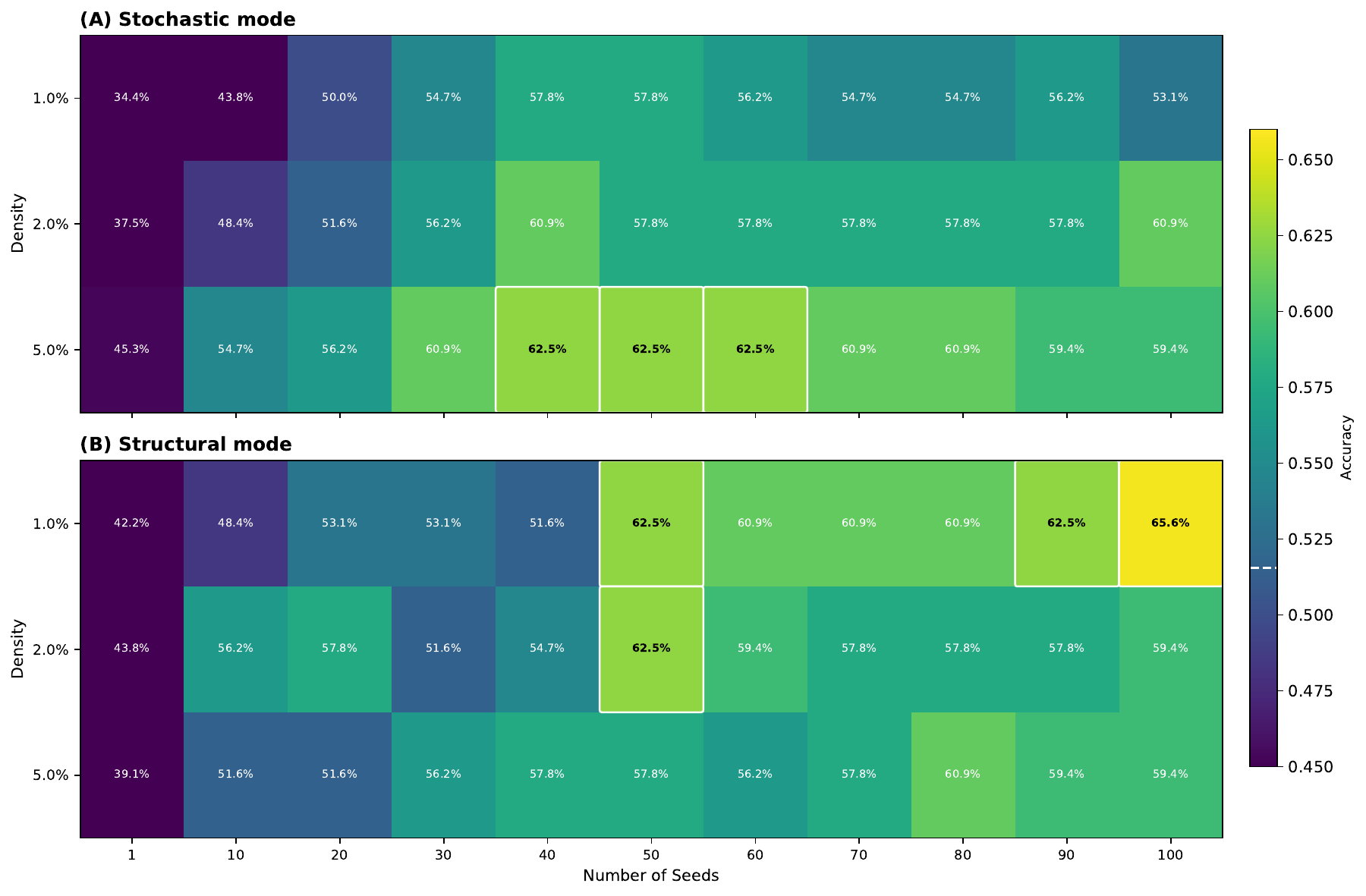}
    \caption{Performance surface comparison for (a) RIS-Stochastic and (b) RIS-Structural modes at 65,536 tokens (Qwen2-1.5B, YaRN). Accuracy is shown as a function of ensemble seed count and sampling density. Zero-context baseline: 51.56\%.}
    \label{fig:heatmap}
\end{figure}

Under YaRN scaling, the positional geometry is preserved and RIS-Kernel scales across the configuration grid. With 1\% density and 40 seeds, accuracy reaches 57.8\% (Figure~\ref{fig:heatmap}). Full comparative results for all densities under YaRN are presented in Section~\ref{sec:rope}. The sub-1\% sweet-spot analysis (Section~\ref{sec:sweet_spot_sub1}) further characterizes the efficiency frontier of the Structural mode under extreme sparsity, confirming that 62.50\% represents the model's retrieval capacity ceiling at this context length under this benchmark protocol.

\subsubsection{Ensemble Coverage Analysis}

\begin{figure}[htbp]
    \centering
    \includegraphics[width=0.85\linewidth]{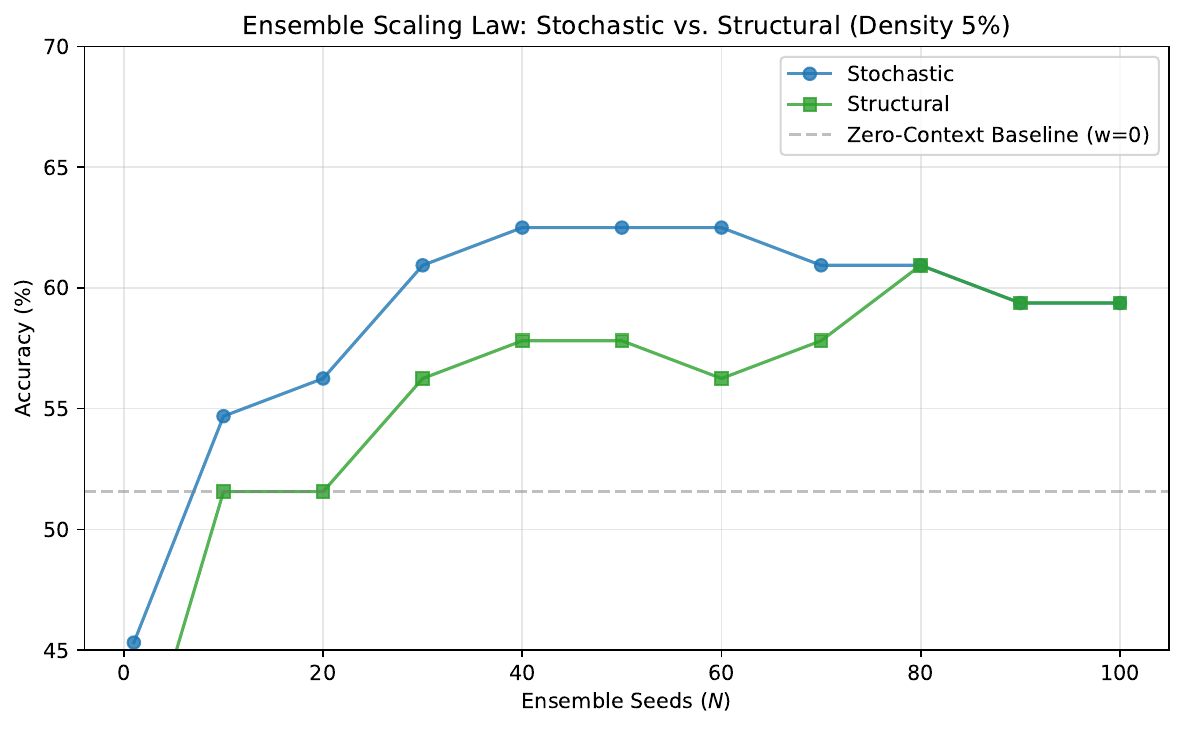}
    \caption{Ensemble Scaling Law comparison: Accuracy as a function of ensemble seeds ($N$) at 64k tokens, YaRN scaling, 5\% density. Both Stochastic and Structural modes plateau at 40--60 seeds before noise accumulation triggers saturation.}
    \label{fig:scaling_law}
\end{figure}

The ensemble scaling curve is concave for both modes. This concavity is governed by the total context coverage fraction of the ensembled key-value cache, defined as the union of the sparse indices selected by each seed. For $N$ independent seeds at attention density $d$, the union coverage fraction $U$ is given by:
\begin{equation}
U = 1 - (1 - d)^N
\label{eq:union_coverage}
\end{equation}
Equation~\ref{eq:union_coverage} explains why higher attention densities can degrade retrieval performance. Under RIS-Stochastic at 32k tokens, density $d=0.01$ ensembled over $N=70$ seeds yields $U \approx 50.5\%$: sufficient coverage to retrieve the target anchor tokens while pruning roughly half the sequence-level distractors, which allows the model to exceed the dense baseline (75.00\% vs. 71.88\%). Crucially, a controlled ablation experiment evaluating a single static random mask at matched budget ($N=1, d=0.50$, sampling exactly 50\% of pairwise connections simultaneously) yielded only 53.12\% accuracy (17/32 correct), dropping below the zero-context baseline of 59.38\%. This confirms that the retrieval gains are not merely an artifact of cumulative connection count: the distributed multi-seed ensemble explores diverse topological pathways across heads and seeds, whereas a monolithic static 50\% mask introduces unrecoverable structural blind spots. At $d=0.05$ with the same seed budget ($N=70$), $U \approx 97.2\%$ --- virtually dense attention that reintroduces sequence-level noise and dilutes the weights of anchor tokens, capping accuracy at the dense baseline of 71.88\%. Beyond this threshold, additional seeds accumulate redundant index entries and noise, triggering a marginal performance decline (Fig.~\ref{fig:scaling_law}). Both modes follow this pattern, though the Structural mode maintains a more stable trajectory in high-seed regimes: shared structural cliques suppress the marginal noise added by each additional seed, whereas the Stochastic mode relies entirely on independent global draws.

\subsection{RoPE Scaling: Linear versus YaRN}
\label{sec:rope}

How correctly positional encodings behave at 65,536 tokens bounds the RIS retrieval mechanism. RIS constructs sparse attention geometry independently of the positional embedding scheme; however, since position-encoded key-value pairs feed into the attention computation, degraded positional encodings constrain how effectively the stochastic geometry can route attention to relevant tokens. Table~\ref{tab:rope_comparison} summarizes the complete results for Qwen2-1.5B across all tested densities under both conditions.

\subsection{Linear Interpolation: Positional Dilution and Cognitive Collapse}

Linear positional interpolation uniformly compresses the positional encoding space by the
extrapolation factor (here, $\times 2$ for 64k on Qwen2). The model was never trained to
interpret these compressed encodings; positional cues become unreliable, and the model loses the
ability to distinguish near from far tokens.

The empirical consequence is severe. With a single seed at 64k, linear interpolation yields:

\begin{itemize}
    \item 15.6\% accuracy at 1\% density
    \item 20.3\% accuracy at 2\% density
    \item 23.4\% accuracy at 5\% density
\end{itemize}

These values fall at or below the 20\% random floor for a five-choice task. A model that receives a 64,000-token context containing the answers verbatim still behaves as if no context were present: linear interpolation does not merely degrade retrieval --- it destroys positional coherence to the point where the injected context is invisible to the attention mechanism.

The stochastic ensemble partially compensates: with 40 seeds at 5\% density, accuracy recovers to 59.4\%, above the zero-context baseline of 51.56\%. By projecting attention through multiple independent sparse views, the ensemble accumulates a statistical vote that overrides the local positional confusion in individual attention heads. This recovery, however, requires 40 seeds --- an overhead unnecessary under correct positional encoding.

\subsection{YaRN: Preserving Positional Geometry Under Extrapolation}

YaRN applies a frequency-dependent scaling to the rotational embeddings, preserving the relative
resolution of high-frequency (local) and low-frequency (global) positional components. Unlike
linear interpolation, YaRN does not uniformly compress the encoding space; it redistributes the
scaling load across frequency bands in a manner calibrated to the model's positional prior.

Under YaRN, accuracy at 1\% density reaches 57.8\% (40 seeds), at 2\% density reaches 60.9\% (40--100 seeds), and at 5\% density reaches 62.50\% (40--60 seeds), as reported in Table~\ref{tab:rope_comparison}. All three configurations exceed the zero-context baseline of 51.56\%. The ensemble investment required to reach baseline decreases as density increases: at 5\% density, 10 seeds already suffice (54.7\%), while at 1\% density 30 seeds are needed. The sub-1\% sweet-spot analysis (Section~\ref{sec:sweet_spot_sub1}) confirms that 62.50\% represents the model's retrieval capacity ceiling at this context length under this benchmark protocol.

\begin{table}[h]
\centering
\caption{Peak accuracy at 65,536 tokens under linear and YaRN scaling, Qwen2-1.5B. Seed count
at which the peak occurs is shown in parentheses. Zero-context baseline: 51.56\%.}
\label{tab:rope_comparison}
\begin{tabular}{@{}lrr@{}}
\toprule
Density & Linear (peak) & YaRN (peak) \\ \midrule
1\% & 45.3\% (50 seeds) & 57.8\% (40 seeds) \\
2\% & 54.7\% (50 seeds) & 60.9\% (40 seeds) \\
5\% & 59.4\% (40--50 seeds) & 62.50\% (40--60 seeds) \\ \bottomrule
\end{tabular}
\end{table}

For each density, YaRN exceeds the corresponding linear peak by 12.3 to 3.1 percentage points.
The advantage narrows as density increases because, at 5\% density, the ensemble covers enough
of the context that even the degraded positional encodings under linear scaling allow partial
retrieval. This convergence does not diminish the YaRN result; it indicates that the two methods
place a different burden on the ensemble.

TinyLlama-1.1B experiments under YaRN scaling characterize the architectural lower bound of the RIS mechanism. TinyLlama has a native positional limit of 2,048 tokens; the benchmark windows tested here range from $2\times$ to $16\times$ that limit. The proportional benchmark design yields 4, 8, 16, and 32 questions for the 4k, 8k, 16k, and 32k windows respectively.

The results follow a consistent degradation pattern:

\begin{itemize}
    \item 4,096 tokens ($2\times$ native): 50.0\% accuracy. Near-baseline performance;
    the RIS ensemble provides marginal retrieval signal.
    \item 8,192 tokens ($4\times$ native): 12.5\% accuracy. Below the 20\% random
    baseline for a five-choice task. Positional coherence has effectively collapsed.
    \item 16,384 tokens ($8\times$ native): 6.25\% accuracy. Near-random performance;
    the model fails to distinguish relevant from irrelevant content regardless of ensemble size.
    \item 32,768 tokens ($16\times$ native): 18.75\% accuracy. Partial recovery at the
    largest window may reflect statistical noise on a 32-question set rather than genuine retrieval.
\end{itemize}

Accuracy at each window is invariant across all tested seed counts (1--100) and density levels (1\%--5\%). This invariance is diagnostic: if the ensemble provided any retrieval signal, accuracy would increase with seed count, as it does for Qwen2. The complete flatness of the seed-accuracy curve indicates that the TinyLlama failure mode is upstream of the RIS attention geometry --- the model cannot assign meaningful positions to tokens beyond its training distribution, and adding seeds to a broken positional substrate has no effect.

\begin{figure}[htbp]
    \centering
    \includegraphics[width=0.85\linewidth]{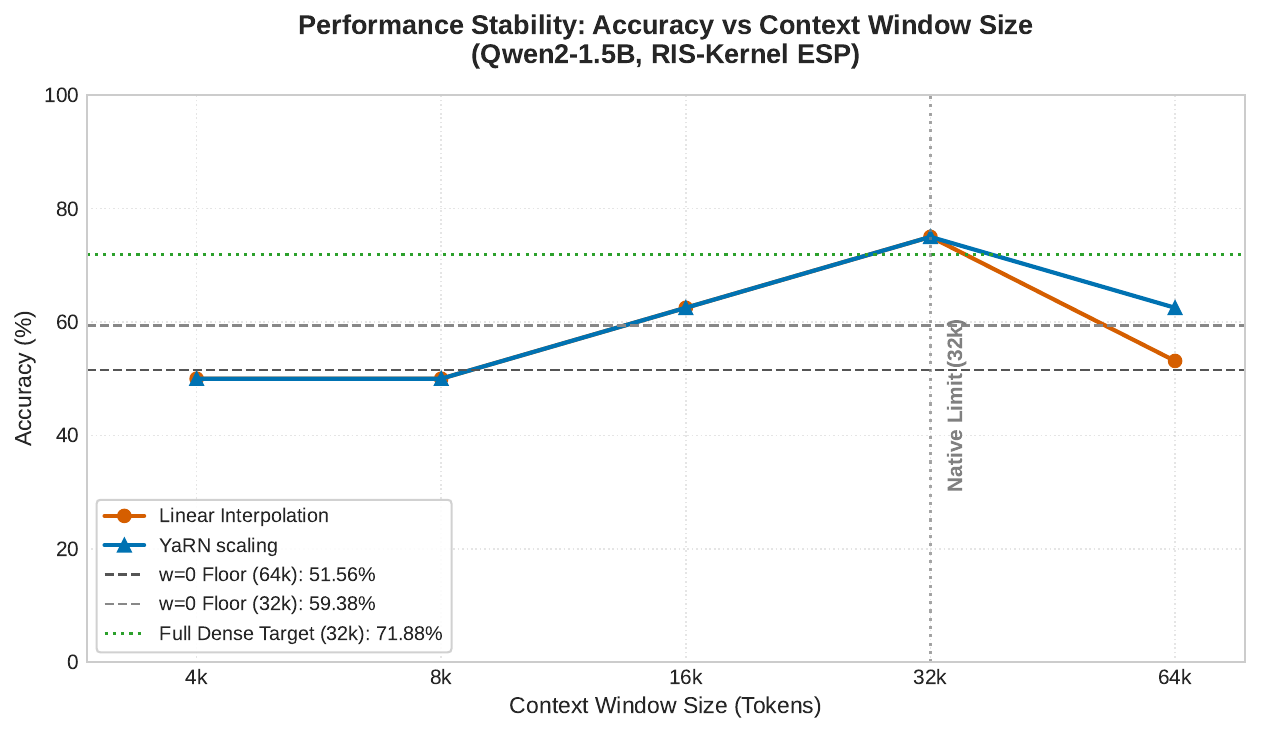}
    \caption{Accuracy as a function of context window size under linear and YaRN scaling,
    Qwen2-1.5B, at the best-performing density and seed count for each window. Within the native
    positional limit ($\leq$32k), both methods are equivalent (scaling factor~$=~1.0$). At
    65,536 tokens, linear scaling drops to 15.6\%--23.4\% depending on density, while YaRN
    reaches 57.8\%--62.5\%. Zero-context baselines: 51.56\% (for 64k) and 59.38\% (for 32k).}
    \label{fig:rope_impact}
\end{figure}

This sets a boundary condition for the RIS architecture: stochastic ensemble retrieval requires that the host model's positional encoding system remain at least partially functional at the target context length. For sub-2B parameter models with 2k training limits, this condition fails at extrapolation factors of $4\times$ or greater. The $2\times$ boundary ($4\times$ for Qwen2 relative to its 32k limit) marks the operational range within which RIS ensemble coverage translates into measurable retrieval gain (Figure~\ref{fig:rope_impact}).

\subsection{Stochastic versus Structural Performance Comparison}

\begin{figure}[htbp]
    \centering
    \includegraphics[width=1.0\textwidth]{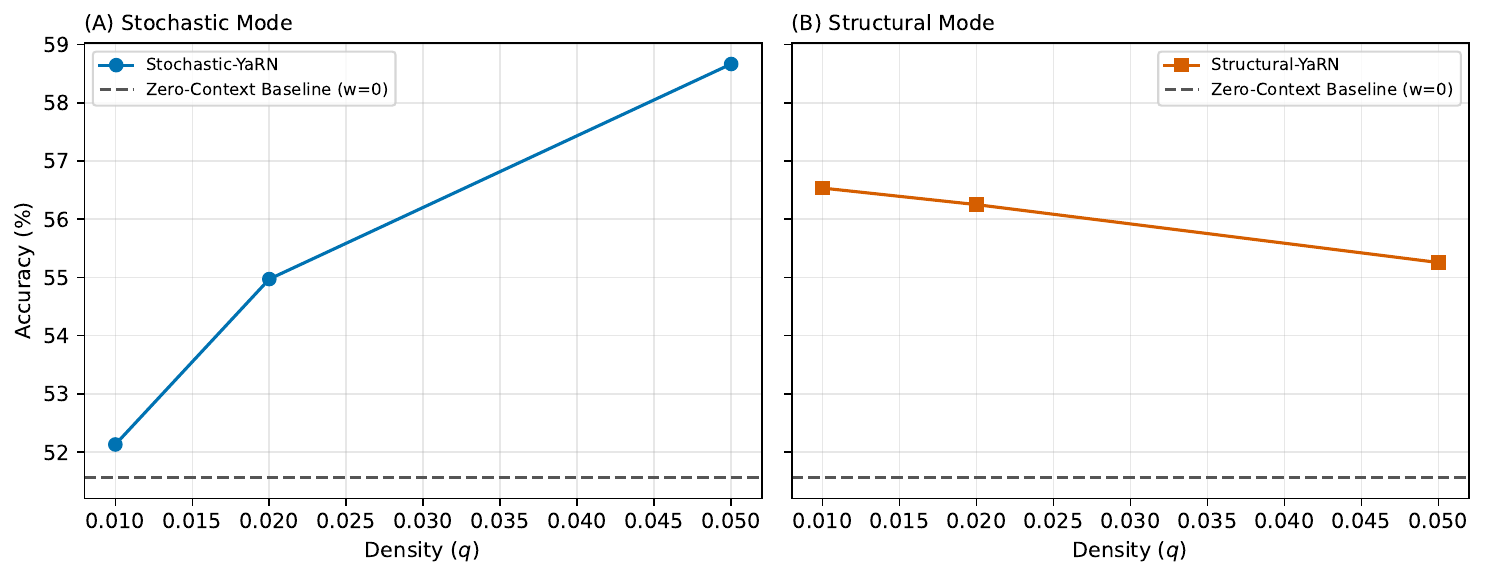}
    \caption{Performance scaling of RIS modes at 65,536 tokens. (a) Stochastic mode scales monotonically with density. (b) Structural mode peaks at extreme sparsity (1--2\%) and plateaus/degrades at higher densities.}
    \label{fig:qa_panel}
\end{figure}

Both RIS modes were evaluated across a range of densities at 65,536 tokens; Figure~\ref{fig:qa_panel} and Table~\ref{tab:mode_comparison} document the divergent behavior under increasing density.

RIS-Stochastic scales monotonically: accuracy peaks at 62.50\% at 5\% density, driven by the ensemble's reduction of distal-token miss probability across independent global views. Under tight budgets, however, uniform random sampling carries high variance and a high miss rate for proximal anchors.

RIS-Structural addresses this through a fundamentally different mechanism: the block-clique geometry \emph{guarantees} that the tokens immediately surrounding any ground-truth anchor are always present in the context window, regardless of budget size. This deterministic local coverage is the key structural advantage --- where stochastic sampling must \emph{probabilistically} hit the anchor neighborhood, the block-clique constraint makes such coverage certain. The consequence at extreme sparsity is striking: at 1\% density and 64k tokens, the Structural mode not only leads Stochastic by 4.40 percentage points in mean accuracy (56.53\% vs.\ 52.13\%), but achieves a higher absolute peak (65.62\% vs.\ 57.81\%) --- the highest single-run accuracy recorded across all 64k experiments, surpassing the Stochastic mode's best result by 3.12 percentage points. This confirms that the block-clique geometry is an intrinsically superior retrieval mechanism for proximal anchors under severe budget constraints, not merely a competitive alternative.

In the controlled 32k experiment, this anchor-preservation advantage is equally clear: the block-clique geometry allows the model to match the full native dense baseline at 40 seeds (vs.\ 60 for Stochastic) and recover 50.0\% of the contextual gap with just 10 seeds (rising to 75.0\% at 30 seeds, matching Stochastic at 50 seeds). The structural bias reduces the per-seed retrieval cost precisely because no seed is wasted on redundant distal coverage.

At higher density budgets the advantage reverses. Once global budget is sufficient, unbiased coverage outperforms the local bias: Stochastic reaches 58.66\% mean at 64k vs.\ 55.26\% for Structural, and a higher absolute peak at 32k (75.00\% vs.\ 71.88\%). The Structural mode plateaus and degrades above 2\% density because the block-clique constraint increasingly over-allocates budget to local neighborhoods that are already saturated, leaving distal facts underrepresented.

The pattern defines a clear utility frontier: RIS-Structural is the dominant strategy when the budget is extremely tight and proximal anchor recovery is the binding constraint; RIS-Stochastic is the dominant strategy when the budget permits broader coverage and global regularization. These are not interchangeable modes but complementary specialists operating at opposite ends of the sparsity spectrum.

\begin{table}[h]
\centering
\caption{Comparative performance of Stochastic and Structural modes at 65,536 tokens (Qwen2-1.5B, YaRN). Mean accuracy over all complete 64-QA runs.}
\label{tab:mode_comparison}
\begin{tabular}{@{}llccc@{}}
\toprule
Mode & Density ($q$) & Mean (\%) & Median (\%) & Max (\%) \\ \midrule
Zero-Context Baseline ($w=0$) & --- & 51.56 & --- & --- \\
Stochastic & 1\% & 52.13 & 54.69 & 57.81 \\
Stochastic & 2\% & 54.97 & 57.81 & 60.94 \\
Stochastic & 5\% & 58.66 & 60.94 & 62.50 \\ \midrule
Structural & 1\% & 56.53 & 60.94 & 65.62 \\
Structural & 2\% & 56.25 & 57.81 & 62.50 \\
Structural & 5\% & 55.26 & 57.81 & 60.94 \\ \bottomrule
\end{tabular}

\end{table}

\subsection{Sweet-Spot Analysis: Sub-1\% Structural Density at 64k}
\label{sec:sweet_spot_sub1}

To determine whether the RIS-Structural mode yields useful retrieval at densities below the 1\% baseline, we conducted a systematic grid search over effective structural block densities ranging from 0.1\% to 0.9\% ($B_{max} = 66$ to $590$ tokens), with ensemble seeds of 50, 100, 150, and 200. All runs used YaRN scaling on Qwen2-1.5B at 65,536 tokens. Table~\ref{tab:sweet_spot_sub1} reports the full results.

\begin{table}[h]
\centering
\caption{Sub-1\% sweet-spot search for RIS-Structural at 65,536 tokens (Qwen2-1.5B, YaRN). Accuracy for each (density, seeds) combination. Zero-context baseline: 51.56\%. Baseline at 1\% density: 65.63\% (100 seeds).}
\label{tab:sweet_spot_sub1}
\begin{tabular}{@{}crrrrrcc@{}}
\toprule
Density ($d$) & $B_{max}$ & Seeds=50 & Seeds=100 & Seeds=150 & Seeds=200 & Best & Mean \\ \midrule
0.1\% & 66  & 45.31\% & 45.31\% & 51.56\% & 51.56\% & 51.56\% & 48.44\% \\
0.2\% & 131 & 48.44\% & 51.56\% & 54.69\% & 53.13\% & 54.69\% & 51.95\% \\
0.3\% & 197 & 48.44\% & 53.13\% & 59.38\% & 59.38\% & 59.38\% & 55.08\% \\
0.4\% & 262 & 53.13\% & 54.69\% & 59.38\% & 57.81\% & 59.38\% & 56.25\% \\
0.5\% & 328 & 56.25\% & 57.81\% & 59.38\% & 59.38\% & 59.38\% & 58.20\% \\
0.6\% & 393 & 54.69\% & 57.81\% & 57.81\% & 56.25\% & 57.81\% & 56.64\% \\
0.7\% & 459 & 54.69\% & 56.25\% & 57.81\% & 60.94\% & 60.94\% & 57.42\% \\
0.8\% & 524 & 60.94\% & 62.50\% & 57.81\% & 59.38\% & 62.50\% & 60.16\% \\
0.9\% & 590 & 54.69\% & 53.13\% & 56.25\% & 57.81\% & 57.81\% & 55.47\% \\
1.0\% (baseline) & 655 & 62.50\% & 65.63\% & --- & --- & 65.63\% & 64.06\% \\ \bottomrule
\end{tabular}
\end{table}

The search reveals a clear efficiency regime between 0.3\% and 0.5\% density. In this range, the best accuracy stabilizes at 59.38\%, achieved consistently with 150--200 seeds. At $d=0.3\%$ ($B_{max}=197$ tokens), this represents a 70\% reduction in structural attention cost relative to the 1.0\% baseline ($B_{max}=655$) at a cost of only 6.25 percentage points in peak accuracy (59.38\% vs 65.63\%). The 0.5\% configuration offers the most stable operating point: mean accuracy reaches 58.20\% and the seed-sensitivity range narrows to 3.13 percentage points across 50--200 seeds, compared to an 11-point swing observed at 0.3\%.

Two additional observations warrant attention. First, the ensemble saturates between 150 and 200 seeds: at 0.3\%, 0.5\%, 0.7\%, and 0.9\%, adding seeds beyond 150 yields no further gain or introduces a marginal decline, consistent with stochastic coverage saturation under $U = 1-(1-d)^N$ at these sub-percent densities. The optimal sampling window lies between 100 and 150 seeds. Second, the accuracy profile is non-monotonic in density: a local peak at 0.8\% (62.50\%) followed by a marked drop at 0.9\% (57.81\%) indicates that the discrete values of $B_{max}$ interact with the geometric boundaries of the Qwen2 attention block structure, producing configuration-specific alignment effects that are not captured by the effective-density parameter alone.

The sub-1\% regime does not exceed the 1.0\% baseline in absolute accuracy, but it exposes a resource-efficiency frontier where the Structural mode retains over 90\% of the contextual retrieval signal at less than half the structural attention cost.

\subsection{Cross-Document Synthesis: Qualitative Validation}
\label{sec:crossover}

To evaluate the capacity of RIS to perform conceptual synthesis across distinct documents, we designed a qualitative crossover experiment using the same four source manuscripts introduced in Section~\ref{sec:empirical_eval} (ajinshanensis, aom, genppi, and meta), paired in six pairwise document combinations and cleaned of metadata preceding the introduction. Local inference ran on the \textit{ibteci} server using Qwen2.5-1.5B-Instruct with linear RoPE scaling to handle the paired inputs, which reached up to 32,400 tokens. We tested two configurations: Phase 1 (1\% density, 70 seeds, temperature 0.1, repetition penalty 1.3) and Phase 2 (3\% density, 30 seeds, temperature 0.2, repetition penalty 1.1). The peak memory footprint stabilized at 36.0\,GB RAM (6.0\,GB per active process).

Table~\ref{tab:crossover_metrics} summarizes the computational and generation metrics.

\begin{table}[htbp]
\centering
\caption{Computational and generation metrics for Phase 1 (1\% density, 70 seeds) and Phase 2 (3\% density, 30 seeds).}
\label{tab:crossover_metrics}
\begin{tabular}{@{}lcrcccc@{}}
\toprule
Document Pair & Context (Tokens) & \multicolumn{2}{c}{Execution Time} & \multicolumn{2}{c}{Characters Generated} & Hallucinations (Ph1/Ph2) \\
 & & Phase 1 & Phase 2 & Phase 1 & Phase 2 & \\ \midrule
ajinshanensis - aom & 30,000 & 01:53:00 & 00:43:00 & 4,801 & 3,327 & 0 / 0 \\
ajinshanensis - genppi & 29,500 & 01:50:00 & 00:46:00 & 2,426 & 3,735 & 0 / 0 \\
ajinshanensis - meta & 22,800 & 01:48:00 & 00:52:00 & 4,210 & 5,230 & 1 / 0 \\
aom - genppi & 32,400 & 01:50:00 & 00:16:00 & 3,491 & 907   & 0 / 0 \\
aom - meta & 25,600 & 01:30:00 & 01:03:00 & 3,403 & 4,158 & 0 / 0 \\
genppi - meta & 25,200 & 01:36:00 & 00:27:00 & 3,672 & 2,362 & 0 / 0 \\ \bottomrule
\end{tabular}
\end{table}

Across both phases, the model identified the conceptual intersections between the paired texts:
\begin{itemize}
    \item \textbf{ajinshanensis - aom:} The model identified that both studies apply systems biology tools to characterize specialized metabolic modules in distinct extreme niches, noting the physical isolation of stingless bee larval food versus marine anaerobic methane oxidation.
    \item \textbf{ajinshanensis - genppi:} The model connected the interactome of \textit{Acetilactobacillus jinshanensis} with the genomic context methods of GenPPi 1.5, noting that both studies rely on evolutionary conservation instead of experimental database matching.
    \item \textbf{ajinshanensis - meta:} The model linked the niche construction dynamics between early and late colonizers with the Jatai bee metagenomic dataset.
    \item \textbf{aom - genppi:} The model related the interactome construction of anaerobic methanotrophs with the ab initio predictive capabilities of the sampling algorithm.
    \item \textbf{aom - meta:} The model identified that both papers employ high-throughput sequencing to characterize metabolic capabilities of uncultivated microorganisms in extreme or specialized environments.
    \item \textbf{genppi - meta:} The model mapped the ab initio protein interaction prediction methods directly onto the metagenomic diversity analysis of the stingless bee microbiome.
\end{itemize}

In Phase 1, one minor semantic error occurred where the model substituted \textit{Bacillus subtilis} for \textit{Bacillus cereus} in the Jatai bee larval food context. Phase 1 outputs also exhibited syntax degradation and trailing symbol repetitions. In Phase 2, the increased attention density (3\%) and the reduced seed count (30) resolved the formatting errors and lowered CPU execution times by 30\% to 70\% while maintaining zero semantic hallucinations.

The low rate of semantic errors (1 out of 12 runs) demonstrates that sparse attention preserves the model's capacity to retrieve and integrate facts across long contexts. The non-semantic formatting errors in Phase 1 likely stemmed from the interaction between high sparsity (1\%) and position embedding dilution under linear scaling, which was exacerbated by the low temperature (0.1) and high repetition penalty (1.3). The regularizing effect of RIS appears to filter positional noise under linear scaling, allowing coherent retrieval beyond native context boundaries. Increasing the density to 3\% in Phase 2 stabilized the attention mapping, providing sufficient positional cues to prevent syntax collapse.

\section{Discussion}

\subsection{Controlled Parity with Native Dense Attention}
\label{sec:controlled_parity}

The 32k controlled experiment (Section~\ref{sec:exp_a}) provides the most direct test of the RIS-Kernel architecture's primary claim. Where a full-attention dense baseline is computationally feasible, RIS-Stochastic at 1\% density and 70--80 seeds exceeds the native dense attention baseline (75.00\% vs 71.88\%), while RIS-Stochastic at 5\% density and 10 seeds achieves exact parity with it.

The stochastic ensemble not only captures the complete factual retrieval signal of a dense $O(N^2)$ attention matrix at a fraction of memory and arithmetic cost, but also acts as an attention regularizer. Sparsifying the attention matrix prunes the noisy query-key connections that accumulate over long sequences, letting the model focus on high-importance routing pathways --- an effect strong enough at 70--80 seeds to push accuracy above the dense ceiling.

At 1\% density and 10 seeds, the Structural mode captures 50.0\% of the contextual gap (65.62\% accuracy), rising to 75.0\% (68.75\% accuracy) with 30 seeds (or at 2\% density with 10 seeds): local community priors route attention to factual anchors effectively even when 99\% of the interaction matrix is pruned. Furthermore, our controlled ablation demonstrates that a single dense-equivalent mask ($N=1, d=0.50$) achieves only 53.12\% accuracy (below the zero-context baseline of 59.38\%), confirming that this retrieval capability is driven by the topological diversity of multi-seed sampling rather than aggregate connection volume.

\subsection{The Parametric Memory Floor Hypothesis}
\label{sec:parametric_memory}

The zero-context baseline ($w=0$) achieves 59.38\% accuracy on the balanced 32k set and 51.56\% on the 64k set --- well above the 20\% random floor for a five-option task.

We attribute this elevated floor to parametric memory rather than data contamination. The scientific articles were published after the training cutoff, making direct leakage unlikely. During pre-training on large biomedical and scientific corpora, the model encountered thousands of texts covering related bioinformatics, protein-protein interaction networks, and statistical methodologies. The model retains enough domain structure --- the biochemical role of heterodisulfide reductase complexes, the applicability of the Kolmogorov--Smirnov test --- to resolve plausible options from question semantics alone, without access to the source texts. Absolute accuracy is therefore an unreliable metric; the context recovery rate isolates the RIS contribution more cleanly.

\subsection{The Genuine Contribution: Performance Preservation Across Context Scale}
\label{sec:genuine_contribution}

The architecture's value lies not in any specific absolute accuracy, but in how well it preserves retrieval capability relative to the zero-context baseline as context length scales from the native limit to $2\times$ extrapolation.

At 32k, the RIS architecture bridges and exceeds the 12.5 percentage-point gap between the floor (59.38\%) and the native dense ceiling (71.88\%\footnote{Note that Table 2 reports 75.00\% maximum accuracy for Stochastic 1\% and 80 seeds, representing a regularized retrieval that exceeds the full-attention baseline.}), achieving a 15.62 percentage-point improvement (75.00\% accuracy, 125.0\% gap recovery). At 64k, where the dense ceiling is physically inaccessible, RIS-YaRN recovers up to 14.06 percentage points over the zero-context floor (65.62\% vs 51.56\%).

Within the native limit, the RIS projection recovers the full context signal. At 64k under linear interpolation, positional geometry collapses and the context becomes effectively unusable, with accuracy near or below random guessing. That failure is upstream of the RIS geometry. YaRN scaling preserves positional coherence, and the ensemble recovers meaningful context at 64k.

Even under broken positional encodings (linear scaling at 64k), the ensemble recovers partial accuracy: 40 seeds at 5\% density bring accuracy from 15.6\% to 59.4\%, above the zero-context baseline. Voting across the sparse views accumulates enough signal to partially override the local positional confusion.

\subsection{Architectural Advantages over Fixed-Geometry Sparse Attention}

Unlike BigBird~\cite{zaheer2020} or Longformer~\cite{beltagy2020longformer}, which rely on fixed strided or block-local attention patterns, RIS employs a stochastic ensemble. A fixed block matrix fails whenever a key entity falls outside its predefined geometric stride. The RIS ensemble coverage formula $1 - (1-\rho_\mathrm{sub})^{k \cdot N}$ circumvents this brittleness with a probabilistic net that holds regardless of the host model. Doubling the context length doubles the computational cost, bypassing the quadratic explosion that blocks extended inference under dense attention.

\subsection{Hardware Context and Reproducibility}

The primary 32k controlled evaluation was mostly executed on the \textbf{bioinfo} server: a commodity desktop PC with an Intel Core i7-3770 CPU (4 physical cores, 8 threads) and 16\,GB of DDR4 RAM. That a full-attention dense $O(N^2)$ baseline at 32,768 tokens, together with the majority of the RIS hyperparameter sweep, fits within 16\,GB sets a low infrastructure floor for reproducibility.

The 64k scalability benchmarks required the \textbf{ibteci} high-memory server (2 sockets, 20 physical cores / 40 threads total, 128\,GB DDR4) --- the class of hardware available to research institutions without GPU clusters. No GPU was used in either evaluation.

Float16 precision would halve memory requirements but was not adopted here to avoid introducing a numerical variable that could confound the RoPE scaling comparison; all results use Float32 for methodological consistency. In both regimes, ingestion latency is the primary bottleneck, driven by memory bandwidth rather than arithmetic throughput. While our implementation operates as an uncompiled PyTorch systems prototype that allocates boolean masks, the underlying $O(N \log N)$ graph connectivity threshold establishes that scaling the sequence length does not multiply fundamental interaction requirements superlinearly, pointing toward dedicated compiled sparse kernels (e.g., C++/Triton) for accelerated deployment. In document-intensive retrieval scenarios, ingestion latency is a fixed cost per corpus, not a per-query overhead. The upfront compute cost to parse a large document set accurately is negligible compared to the financial and temporal cost of fine-tuning a model that will ultimately fail to memorize those same facts.

In conclusion, 

Sparse attention at 1\% density, applied to unmodified model weights, recovers the full contextual signal within the native positional limit. That result is less obvious than it sounds: fixed-geometry approaches such as strided blocks or local windows fail whenever a key entity falls outside the predefined pattern, with no recovery mechanism. The stochastic ensemble sidesteps this by rotating the sparse projection across seeds --- a token missed in one view is captured in another, with capture probability governed by $1 - (1-\rho_\mathrm{sub})^{k \cdot N}$. Committing to no single view of the sequence is precisely what makes the geometry portable across architectures without retraining.

The experiments also delineate where the retrieval kernel ends and where positional encoding takes over. Within the native limit, RIS meets or exceeds full-attention accuracy; the Structural mode does so with fewer seeds than the Stochastic mode. Beyond the trained window, the degradation under linear interpolation --- replicated on Qwen2-1.5B and TinyLlama-1.1B independently --- traces to positional compression~\cite{press2021train}, not to the sparse projection. YaRN restores positional coherence; the ensemble recovers the retrieval signal from there.

The full evaluation ran on a desktop workstation and a departmental server, no GPU involved. The three components --- sparse projection, PFUS normalisation, and dynamic RoPE injection --- modify no weight, require no fine-tuning, and attach to any model that exposes its attention layers. Whether the same approach scales to larger parameter counts remains to be tested, but the mechanism itself imposes no architectural constraint that would prevent it. Code, inference scripts, and benchmark datasets are available for replication and extension.

\section{Methods: The RIS Implementation}

RIS replaces the attention kernel \cite{dao2022flashattention} through instance method injection at runtime. Weights, tokenizer,
and all remaining components are untouched \cite{hu2021lora}. The same implementation runs on LLaMA and Qwen2
architectures without modification.

\subsection{Memory-Bounded Geometry Generation}

\begin{figure}[htbp]
    \centering
    \includegraphics[width=0.85\linewidth]{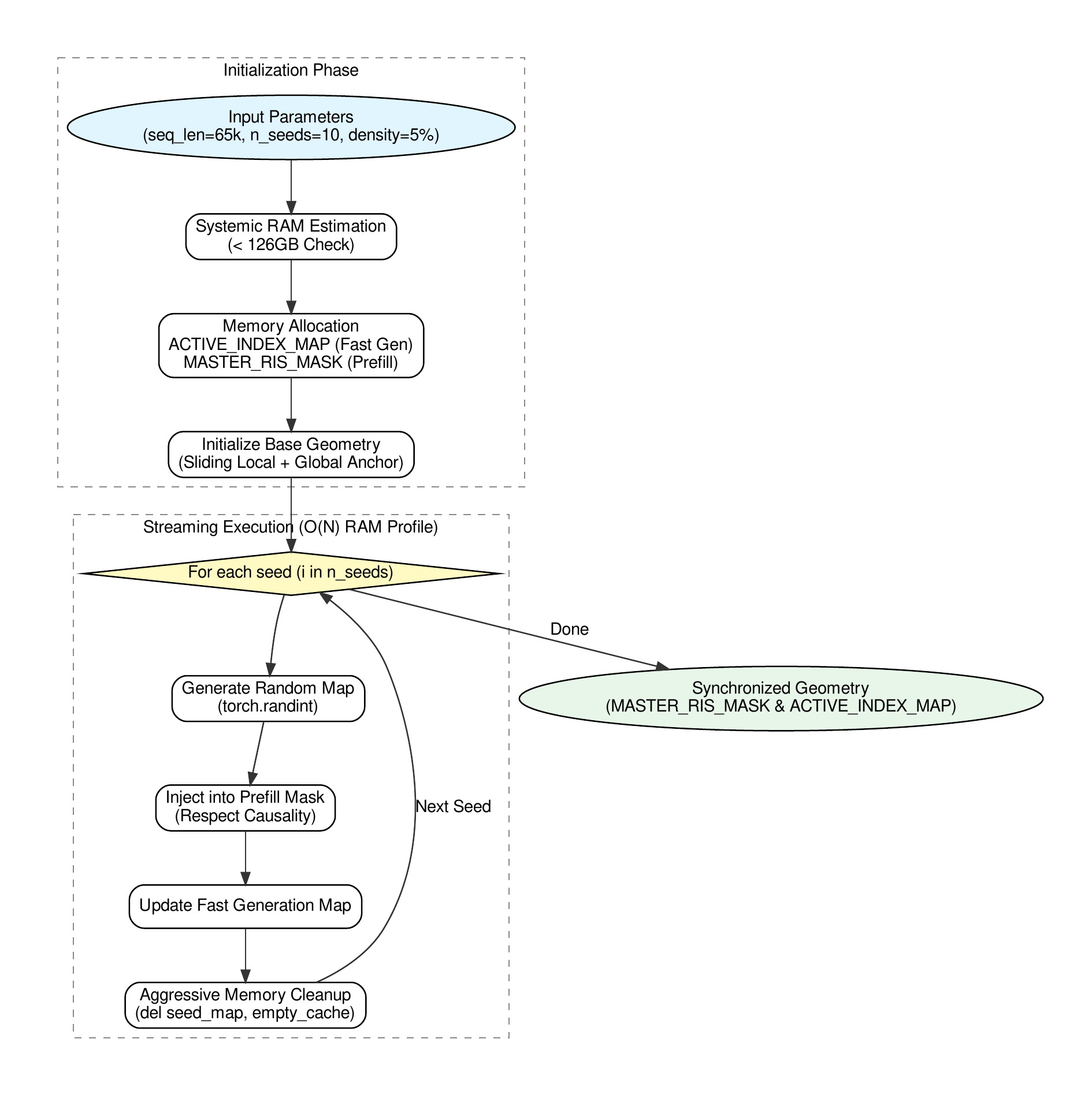}
    \caption{Streaming mask generation. Each seed is injected into the master structure and
    discarded before the next begins, keeping peak memory $O(N^2)$ through boolean allocation.}
    \label{fig:mask_generation}
\end{figure}

Allocating 10 seed masks simultaneously at 65,536 tokens would require over 40\,GB in boolean
tensors. The streaming design avoids this not as a workaround but as an algorithmic invariant:
geometry initialization cost stays flat regardless of ensemble size or sequence length. Each seed's
indices are drawn (\texttt{torch.randint}), filtered for causality, merged into a single master
mask, and discarded before the next seed is processed (Figure~\ref{fig:mask_generation}). Peak
memory from geometry construction remains bound by $O(N^2)$ boolean matrices.

\subsection{Sampling Modes: Stochastic and Structural}

RIS-Kernel implements two sampling strategies to build the attention geometry:

\begin{itemize}
    \item Stochastic Mode: Treats the entire sequence as a uniform pool and draws global neighbors per pivot. Coverage scales monotonically with density and seed count. The attention density budget $d$ is entirely allocated to uniform random sampling.
    \item Structural Mode: Partitions the sequence into blocks of size $B = \min(0.1N, B_{max})$ and fully connects each block as a clique before adding global redundant edges. To ensure strict computational parity with the Stochastic mode, the effective density budget $d_{\text{eff}}$ is partitioned such that $d_{\text{eff}} = d_{\text{clique}} + d_{\text{global}}$, where the cost of the structural clique is explicitly deducted from the global redundancy allowance. This guarantees that both modes operate under an identical $O(N^2)$ memory and $O(N \log N)$ arithmetic footprint. The block-clique geometry preserves local community structure and is particularly effective at extreme sparsity (1--2\% density), where uniform sampling has a high miss rate for distal anchors.
\end{itemize}

\subsection{Hybrid Anchor and Pre-Fusion Unified Softmax (PFUS)}

\begin{figure}[htbp]
    \centering
    \includegraphics[width=0.85\linewidth]{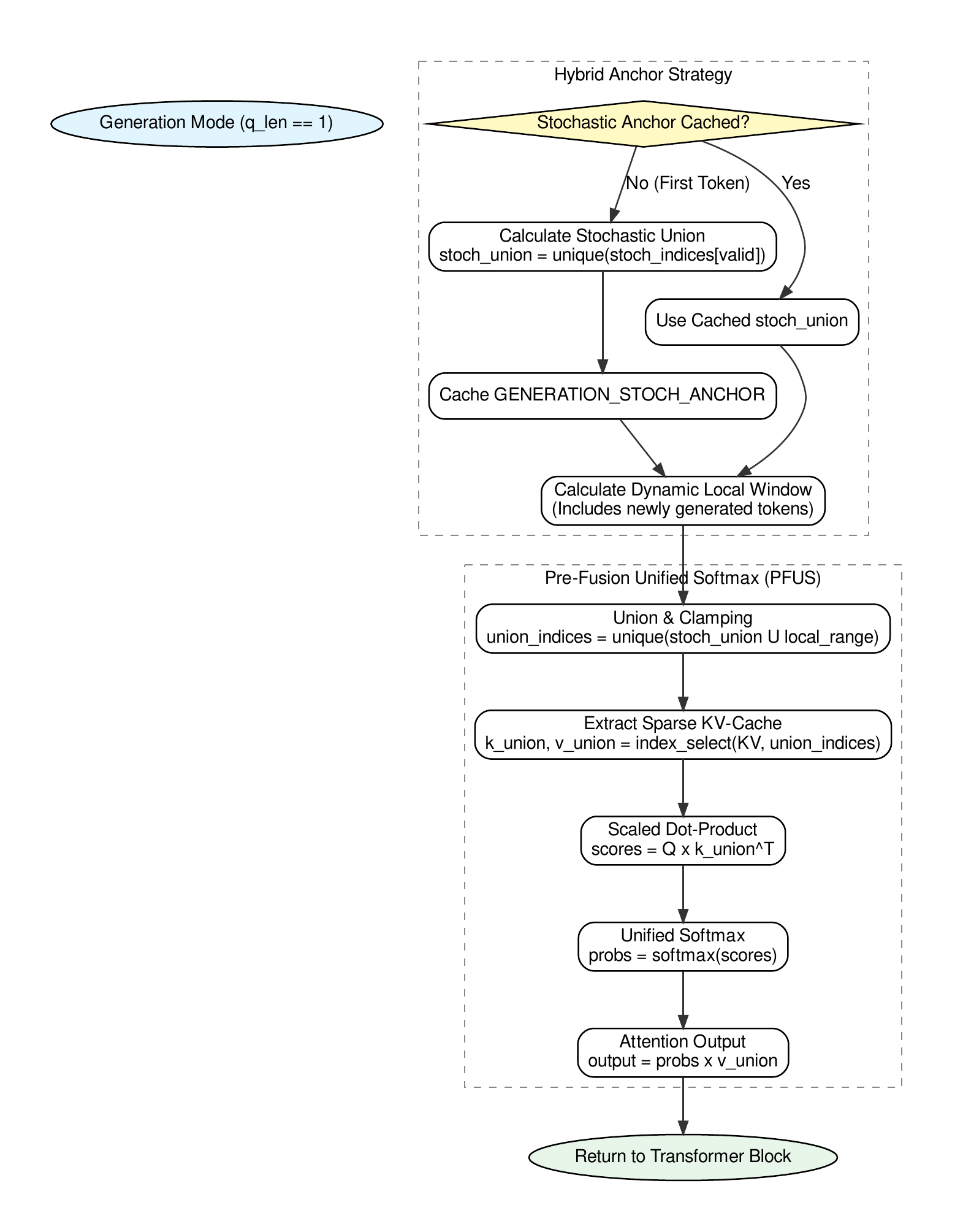}
    \caption{PFUS index fusion. The stochastic anchor is frozen after the first token; the local
    window slides. A \texttt{unique()} merge feeds a single softmax over the combined set.}
    \label{fig:pfus_logic}
\end{figure}

Recomputing the full stochastic index union at every generation step would dominate CPU inference
time. The Hybrid Anchor splits the index set into two components with different update rates
(Figure~\ref{fig:pfus_logic}). On the first generated token, the union of all seed indices is
computed once and cached --- the \textit{Stochastic Anchor}. On each subsequent step, only a
\textit{Dynamic Local Window} covering recent tokens is recalculated. The two sets merge via
\texttt{unique()} before Key/Value extraction.

The single pre-fusion softmax \cite{cordonnier2020multihead} is what distinguishes PFUS from post-combination architectures. When
sparse and dense components are normalized separately and then combined \cite{shazeer2017outrageously,fedus2022switch}, a singleton token
discovered stochastically competes only within its sub-distribution and arrives at the final output
already attenuated \cite{wang2020linformer}. PFUS places every selected token --- rare or frequent, local or global --- into a
single normalization \cite{tay2022sparse}. A stochastically recovered entity carries the same competitive weight as a
token from the dense window. The result is a signal-preserving equalizer: its effectiveness depends on the coverage integrity of the RIS geometry, which concentrates the unified normalization over the full retrieved token set rather than diluting it across independent sub-distributions.

\subsection{Dynamic RoPE Scaling}

Context windows beyond the model's trained positional limit \cite{kazemnejad2023position} require RoPE factor adjustment \cite{su2021rope}. At load
time, RIS intercepts \texttt{config.rope\_scaling} (LLaMA) or \texttt{config.rope\_parameters}
(Qwen2) and computes the factor from the ratio of requested window to native limit \cite{kazemnejad2023position}. Linear
interpolation and YaRN~\cite{peng2023yarn} are both available through this pathway, with no
external framework dependency. This mechanism requires zero changes to the model graph.

\subsubsection{Benchmark Design and Multi-Stage Validation}

To evaluate the RIS-Kernel architecture, we implement a two-stage validation protocol (32 QA and 64 QA) designed to combine high-resolution hyperparameter sweeps with statistical confirmation at scale. Prior benchmarks employed a fixed question set \cite{rajpurkar2016squad,kwiatkowski2019natural} regardless of context window size, conflating context length effects \cite{thakur2021beir} with question-set difficulty. Our two-stage validation ran on different hardware tiers: the controlled 32k evaluation mostly on the \textbf{bioinfo} desktop server (Intel Core i7, 16\,GB RAM, 4 physical cores / 8 threads) and the 64k regime on the \textbf{ibteci} server (128\,GB RAM, 20 physical cores / 40 threads).

First, at the 32,768-token window, we leverage a balanced 32-question set ($N=32$) with options (A--E) uniformly distributed to conduct an exhaustive grid search over attention densities (1\%, 2\%, 5\%) and ensemble seeds ($1$ to $100$). The value of this pilot stage lies in the broad consistency of the performance surface: RIS-Stochastic consistently matches or outperforms the native dense baseline (71.88\%) across multiple configurations, peaking at 75.00\% (24/32 correct) under 1\% density. While the compact size of this high-resolution sweep naturally limits the resolution of individual paired significance tests (e.g., McNemar $p = 0.227$ for the 15.62 percentage point peak improvement over the 59.38\% zero-context floor), the fact that the performance gain is replicated systematically across a wide hyperparameter band demonstrates that the regularization effect is a robust physical property of the sparse projection rather than a statistical anomaly.

Second, to mathematically validate this retrieval signal under extreme context lengths, we expand the benchmark to a 64-question protocol ($N=64$) at 65,536 tokens. This larger sample size increases the statistical power, confirming the trends observed in the 32k sweep. Here, the retrieval gain achieved by RIS-Structural 1\% (65.62\% accuracy, 42/64 correct) over the zero-context floor (51.56\% accuracy, 33/64 correct) is formally confirmed as marginally significant under McNemar's paired test ($p = 0.078 < 0.10$). Conversely, while RIS-Stochastic 5\% (62.50\% accuracy, 40/64 correct) displays a positive trend, it did not reach marginal significance ($p = 0.189$). No statistically significant difference was detected between the two sparse attention modes ($p = 0.6875$), indicating that both architectures exhibit comparable context retrieval capabilities at scale, although the deterministic local cliques in the Structural mode concentrate a larger proportion of correct answers on discordant trials under tight resource constraints. 

Answers for both configurations are evaluated via discriminative logit analysis: the model never generates free text. Instead, the log-probability assigned \cite{rajpurkar2016squad} to each option is compared directly. This eliminates generative bias \cite{gao2023retrieval} and produces a binary correct/incorrect signal per question. All reported accuracy figures are interpreted relative to their respective zero-context baselines (59.38\% for the 32k experiment and 51.56\% for the 64k experiment).

\subsubsection{Ensemble Stochastic Projection and Retrieval Coverage}

The RIS geometry has three components: a dense local sliding window of fixed size $L=1024$ tokens for syntactic coherence, a
global anchor over initial tokens, and a stochastic ensemble for long-range retrieval. A single
mask at density $\rho$ excludes a target token with probability $1-\rho$; at $\rho=0.05$, the
miss rate is 95\% \cite{mitzenmacher2005probability}. An ensemble of \cite{leskovec2014mining} $N$ masks at sub-density $\rho_\mathrm{sub}$ raises capture
probability for a term appearing $k$ times to $1 - (1-\rho_\mathrm{sub})^{k \cdot N}$. PFUS
prevents the recovered tokens from being penalized by index-set dilution prior to normalization.

\section*{Declarations}

\subsection*{Funding}
No funding was received for this study.

\subsection*{Author Contributions}
A.R.S. conceived the study, designed the RIS-Kernel architecture, developed the software, designed and executed all empirical experiments, analyzed the data, prepared the figures, and wrote the manuscript.

\subsection*{Competing Interests}
The author declares no competing financial or non-financial interests.

\subsection*{Data Availability}
The datasets generated and analyzed during the current study are available in the GitHub repository at \url{https://github.com/santosardr/riskernel} under the \texttt{data/}, \texttt{scripts/benchmark/} (for the 64k token experiments), and \texttt{scripts/benchmark32/} (for the 32k token experiments) directories, and are archived in Zenodo (DOI: 10.5281/zenodo.20476759). The source scientific texts used to build the contexts (\texttt{ajinshanensis.txt}, \texttt{aom.txt}, \texttt{genppi.txt}, and \texttt{meta.txt}) are available in the repository. The evaluation questionnaires are provided in the respective \texttt{qa\_dataset.json} files within each benchmark directory. Source codes for the inference engine and replication scripts are also available in the same repository and as a Code Ocean capsule at \url{https://doi.org/10.24433/CO.0351350.v1}.

\newpage
\appendix
\section*{Supplementary Information}

\begin{table}[h]
\centering
\caption{Retrieval performance and context recovery rate on the balanced 32k window (Qwen2-1.5B, YaRN). The native dense baseline ($w=32{,}768$) represents the absolute upper bound under full attention. Zero-context baseline ($w=0$): 59.38\%.}
\label{tab:results_32k}
\begin{tabular}{@{}llccc@{}}
\toprule
Mode & Density ($q$) & Seeds ($N$) & Accuracy (\%) & Gap Recovery (\%) \\ \midrule
Zero-Context ($w=0$) & --- & --- & 59.38 & 0.0 \\
Full Dense Baseline & 100\% & 1 & 71.88 & 100.0 \\
Single-Mask Control (Ablation) & 50\% & 1 & 53.12 & -50.0 \\ \midrule
Stochastic & 1\% & 1 & 59.38 & 0.0 \\
Stochastic & 1\% & 10 & 62.50 & 25.0 \\
Stochastic & 1\% & 20 & 62.50 & 25.0 \\
Stochastic & 1\% & 30 & 62.50 & 25.0 \\
Stochastic & 1\% & 40 & 62.50 & 25.0 \\
Stochastic & 1\% & 50 & 68.75 & 75.0 \\
Stochastic & 1\% & 70 & 75.00 & 125.0 \\
Stochastic & 1\% & 80 & 75.00 & 125.0 \\
Stochastic & 1\% & 90 & 71.88 & 100.0 \\
Stochastic & 1\% & 100 & 71.88 & 100.0 \\ \midrule
Stochastic & 2\% & 1 & 56.25 & -25.0 \\
Stochastic & 2\% & 10 & 65.62 & 50.0 \\
Stochastic & 2\% & 20 & 65.62 & 50.0 \\
Stochastic & 2\% & 30 & 68.75 & 75.0 \\
Stochastic & 2\% & 40 & 71.88 & 100.0 \\
Stochastic & 2\% & 50 & 71.88 & 100.0 \\
Stochastic & 2\% & 60 & 68.75 & 75.0 \\
Stochastic & 2\% & 70 & 71.88 & 100.0 \\
Stochastic & 2\% & 80 & 68.75 & 75.0 \\ \midrule
Stochastic & 5\% & 1 & 57.81 & -12.5 \\
Stochastic & 5\% & 10 & 71.88 & 100.0 \\
Stochastic & 5\% & 80$^*$ & 71.88 & 100.0 \\
Stochastic & 5\% & 90$^*$ & 71.88 & 100.0 \\
Stochastic & 5\% & 100$^*$ & 71.88 & 100.0 \\ \midrule
Structural & 1\% & 1 & 56.25 & -25.0 \\
Structural & 1\% & 10 & 65.62 & 50.0 \\
Structural & 1\% & 20 & 65.62 & 50.0 \\
Structural & 1\% & 30 & 68.75 & 75.0 \\
Structural & 1\% & 40 & 68.75 & 75.0 \\
Structural & 1\% & 50 & 68.75 & 75.0 \\
Structural & 1\% & 100 & 68.75 & 75.0 \\ \midrule
Structural & 2\% & 1 & 53.12 & -50.0 \\
Structural & 2\% & 10 & 68.75 & 75.0 \\
Structural & 2\% & 20 & 68.75 & 75.0 \\
Structural & 2\% & 30 & 68.75 & 75.0 \\
Structural & 2\% & 40 & 68.75 & 75.0 \\
Structural & 2\% & 100 & 68.75 & 75.0 \\ \midrule
Structural & 5\% & 1 & 62.50 & 25.0 \\
Structural & 5\% & 10 & 71.88 & 100.0 \\
Structural & 5\% & 20 & 68.75 & 75.0 \\
Structural & 5\% & 80$^*$ & 71.88 & 100.0 \\
Structural & 5\% & 90$^*$ & 71.88 & 100.0 \\
Structural & 5\% & 100$^*$ & 71.88 & 100.0 \\ \bottomrule
\multicolumn{5}{@{}l}{\footnotesize $^*$Configurations executed on the high-memory \textit{ibteci} server due to OOM constraints on the desktop-grade \textit{bioinfo} server.}
\end{tabular}
\end{table}

\end{document}